# Utility-Based Control for Computer Vision

Tod S. Levitt*, Thomas O. Binford**, Gil J. Ettinger*, and Patrice Gelband*

*Advanced Decision Systems, Mountain View, California
**Stanford University, Stanford, California

## 1. INTRODUCTION

Several key issues arise in implementing recognition in terms of Bayesian networks. Computational efficiency is a driving force. Perceptual networks are very deep, typically fifteen levels of structure. Images are very wide, e.g. an unspecified number of edges may appear anywhere in an image 512 x 512 or larger. For efficiency, we dynamically instantiate hypotheses of observed objects. The network is not fixed, it is only partially instantiated. Hypothesis generation and indexing are important, but they are not considered here [Nevatia 73], [Ettinger 88]. This work is aimed at near-term implementation with parallel computation in a radar surveillance system, ADRIES [Levitt 88], [Franklin 88] and a system for industrial part recognition, SUCCESSOR [Binford 88].

For many applications, vision must be faster to be practical and so efficiently controlling the machine vision process is critical. Perceptual operators may scan megapixels and may require minutes of computation time. It is necessary to avoid unnecessary sensor actions and computation. Parallel computation is available at several levels of processor capability. The potential for parallel, distributed computation for high-level vision means distributing non-homogeneous computations. This paper addresses the problem of control in machine vision systems based on Bayesian probability models.

We separate control and inference to extend the previous work [Binford 87] to maximize utility instead of probability. Maximizing utility allows adopting perceptual strategies for efficient information gathering with sensors and analysis of sensor data. Results of controlling machine vision via utility to recognize military situations are presented in this paper. Future work extends this to industrial part recognition for SUCCESSOR.

## 2. BAYESIAN NETWORK FOR EVIDENTIAL ACCRUAL

The relationship between models, hypotheses, and decisions is pictured in Figure 2-1. Models represent physical objects in the world, such as military units, formations, industrial parts,

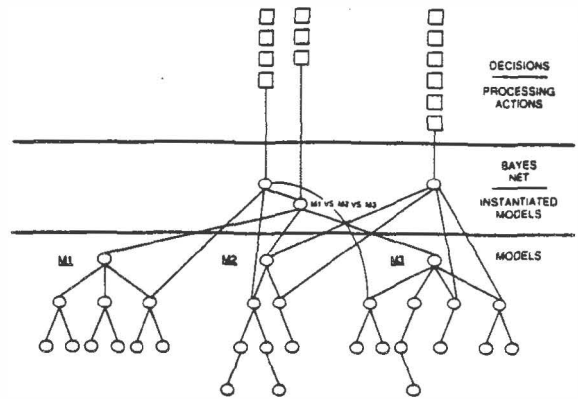

Figure 2-1: Model-Hypothesis-Decision Relationships

245

components of parts, and attributes such as color, reflectivity, etc. As such, we view our models as causal; i.e., a physical object is viewed as "causing" its component sub-parts.

Object models are physical models; their geometry is represented by part/whole graphs and by interlocking taxonomic graphs. Figure 2-2 shows two part-of slices of the (taxonomic) is-a hierarchy for a model of a military brigade of the evil empire of Mordor. The part-of hierarchy corresponds to (physical) military sub-units. The is-a hierarchy is obtained by taking the common set of unit type and formation constraints for military units that can be confused based on uncertain observations. For example, if we are too far away to distinguish steam engines from catapults, we might still recognize them as vehicles, and be uncertain as to whether we are observing a Catapult Battalion or a Steam Engine Team. For military units, the models of military organization predict the existence and location of other sub-units, given the observation of another.

In optical part recognition for manufacturing, we represent objects as part-of hierarchies based on generalized cylinder volume primitives. Object models are recursively broken up into joints composed of parts; those parts may in turn be broken into sub-joints and sub-parts, or they may be primitive. Joints are relationships between parts, incorporating observable effects of joining parts. Such a hierarchy forms a directed acyclic graph (DAG), where nodes are parts or relations and arcs indicate part-of relationships. Generalized cylinders (GCs) are defined by a cross section swept along a space curve, the axis, under a sweeping transformation [Ponce 88]. Compound object models are DAGS of primitives represented in a simple modelling language. Models also include material modeling of optical properties, i.e., reflectives, specularities, and color [Healey 87, 88]. Figure 2-3a shows an elbow without threads. Figure 2-3b shows the line drawing of the elbow without hidden line suppression to show its subparts.

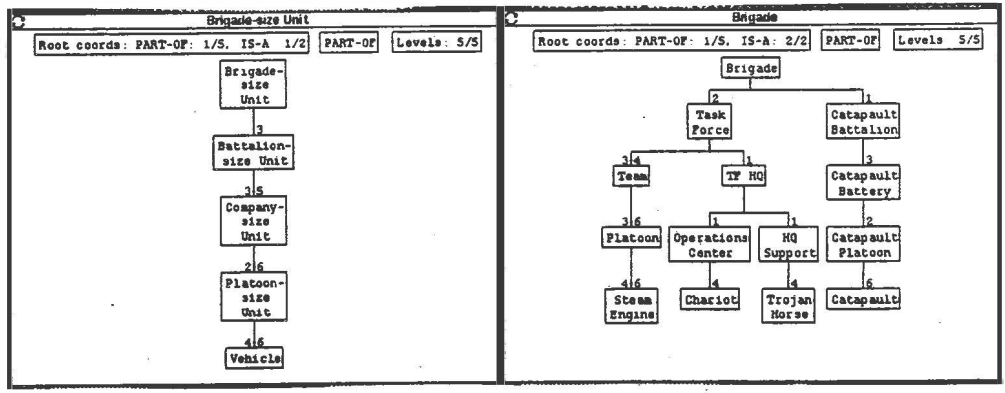

Figure 2-2: Brigade Model

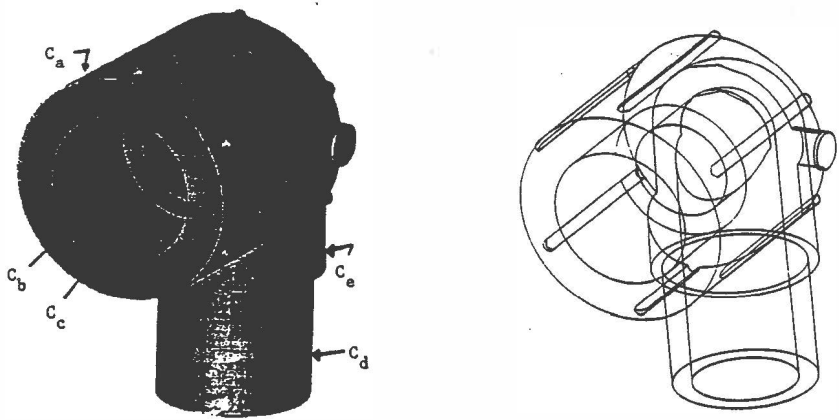

Figure 2-3: a. Elbow without Threads    b. Line Drawing of Elbow without Threads

246

At each node of the Bayes net, there is a probability distribution over the set of mutually exclusive and exhaustive possible interpretations of the visual evidence accrued to that level in the hierarchy. A node is a set of hypotheses, e.g., catapult-battalion vs. task-force vs. non-military-unit, or t-joint versus elbow-joint versus non-joint. Although they do not have to be simultaneously instantiated, the possible links between nodes are hard-wired, a priori, by the models of objects and relationships, and the criteria for node instantiation that determine which pieces of evidence can generate conflicting hypotheses. Each alternative hypothesis at a node contributes some probability to the truth of an alternative hypothesis at a parent node (e.g., the part supports the existence of the whole) and also contributes to the truth of supporting children. When new evidence appears at a node, it is assimilated and appropriate versions of that evidence are propagated along all other links entering or exiting the node. We use the propagation algorithms of Pearl [Pearl 86], [Binford 87].

As we dynamically create the Bayes net at runtime, node instantiation is guided by the a priori models of objects, the evidence of their components, and their relationships. System control alternates between examination of the instantiated Bayes nodes, comparing against the models, and choosing what actions to take to grow the net, which is equivalent to seeing more structure in the world. Thus, inference proceeds by choosing actions from the model space that create new nodes and arcs in the Bayes net. All possible chains of inference that the system can perform are specified a priori in the model-base. This feature clearly distinguishes inference from control. Control chooses actions and allocates them over available processors, and returns results to the inference. Inference uses the existing Bayes net, the current results of actions (i.e., the collected evidence) generates Bayes nodes and arcs, propagates probabilities over the net, and accumulates the selectable actions for examination by control. In this approach, it is impossible for the system to reason circularly, as all instantiated chains of inference must be supported by evidence in a manner consistent with the model-base.

The prioritization and selection of actions can be viewed as a decision-making procedure. By representing the selection of actions at a single Bayes-node as a single decision, we create an influence diagram [Shachter 86] with the property that severing any decision node from the diagram leaves the Bayes-net intact. Figure 2-4 illustrates this design. This allows us to construct control algorithms over the influence diagram where evidence accrual in the Bayes-net, and decisions of actions to execute, appear as modular operations.

## 3. UTILITY FOR EVIDENCE-GATHERING ACTIONS

Our approach to selecting actions by utility theory is to compute the estimated value and cost of each action, then maximize value constrained by a bound on the total cost. We define cost of an action as the average processing time for the action. If the action is an algorithm that can be performed on different processors with radically different computation times, we can model this as two different actions.

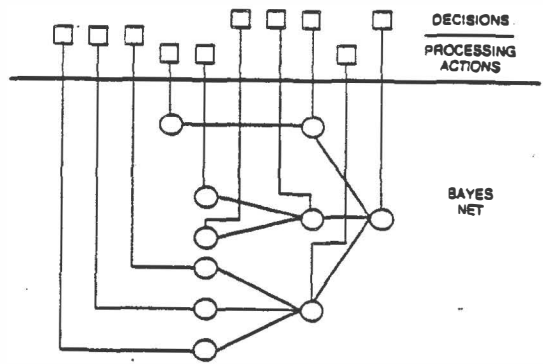

Figure 2-4: Separable Influence Diagram

The computation of value is performed hierarchically over the Bayes net, where hierarchy is the hierarchy inherent in the model space. That is, we view computing the value of an action, $A$, at the child-hypothesis, $H_k$, as the increment in evidential value achieved at the parent hypothesis, $P$. We define the value, $V$, as

$$V \text{ (Child, Action)} = V(H_k, A)$$
$$= \sum_{\text{Parents}} | p(\text{Parent} | H_k, A) - p(\text{Parent} | H_k) | \cdot V(\text{Parent})$$

Thus, we can begin at the top level of the model hierarchy and assign values to recognizing, for example, the various military units or industrial parts. We then, recursively compute the value at each child, or sub-part, down the model hierarchy. In the instantiated Bayes-net, this computation is proportional to the number of instantiated levels in the hierarchy. For example, in trying to confirm or deny the presence of a task force, we can assign a value of .8 to $H_1 = $ task force, .1 to $H_2 = $ catapult battalion and .1 to $H_3 = $ other. These are the objects in the goal Bayes node. The actions include "search for sister sub-unit", "get closer observation of vehicle types", and "adjust match of formation based on adaptation to underlying terrain".

If we have a set of child hypotheses, $H_k$, at Bayes-node $N$, then the value of taking action $A$ at node $N$ is defined as

$$V(N, A) = \sum_k V(H_k, A)$$

If action $A$ has cost $t_A$, we maximize expected value

$$\sum_{(N,A)} \chi_A V(N, A)$$

subject to the constraint that total cost is bounded by $T$:

$$\sum_A \chi_A t_A \leq T$$

where

$$\chi_A = \begin{cases} 1 & \text{if we perform A} \\ 0 & \text{if we do not perform A} \end{cases}$$

and $T$ is the maximum allowable processing time. For any fixed $T$, we produce an equivalence class of plans of actions to be performed, and the results of executing these plans are recognized objects with probabilities. We vary $T$ to obtain the desired level of performance, i.e., we generate sets of plans for each value of $T$. We typically choose the minimum $T$ for a desired probability of recognition.

There is an implicit assumption in this approach that all executable actions are represented in the model space a priori, and that values are calculated to account for continuous ranges of values for individual pieces of evidence. For example, executing a procedure to infer the curvature of a part may depend upon hypothesizing and testing against possible curvatures. We use the expected value of a quasi-invariant measure given its observation as in [Binford 87].

We then use an integer optimization procedure to select over the possible sets of actions executable in the allowed time. The required algorithm for maximizing utility must solve the classic "knapsack" problem. The knapsack problem is to

$$\text{maximize } \sum_i V_i z_i$$

$$\text{subject to the constraints } \sum_i z_i t_i \leq T \text{ and } z_i = 0 \text{ or } 1.$$

The $z_i$ are evidence gathering actions, $V_i$ is the value of a given action, and $t_i$ is the time to perform the action. So the problem is to maximize value by choosing which actions to perform (corresponding to $z_i = 1$) and which not to perform (corresponding to $z_i = 0$) while staying within the time limit.

The knapsack problem is an NP hard problem [Garey 79]. Because we expect to be dealing with on the order of 100 actions, it is infeasible to solve the problem exactly. Therefore, we use an algorithm that finds an approximate solution. Specify the desired accuracy $\epsilon$, and it finds a solution satisfying $\frac{(P' - P)}{P'} < \epsilon$, where $P'$ is the total value of the optimal solution. The algorithm has time complexity $O(N \ln N) + O\left(\left(\frac{3}{\epsilon}\right)^{2 * N}\right)$ and space complexity $O(N) + O\left(\left(\frac{3}{\epsilon}\right)^3\right)$ where $N$ is the number of actions.

We now return to computing

$$p \; (\text{Parent} \,|\, \text{Child, Action}).$$

In general, the increase in belief in the parent depends on the results of computations performed in the action, which can, in turn, depend on many other results of processing at other nodes corresponding to sub- and super-hypotheses. We apply Bayes rule.

$$p \; (\text{Parent} \,|\, \text{Child, Action}) = \frac{p \; (\text{Child, Action} \,|\, \text{Parent}) \; p \; (\text{Parent})}{p \; (\text{Child, Action})}$$

and note that $p\,(\text{Parent})$ is known at runtime. When a Bayes node is already instantiated, and the $p\,(\text{Child, Action})$ can be interpreted as the accuracy with which the results of the action can be measured, given the state of the child. For example, if the child is a boundary of a generalized cylinder of the parent and the action is a curvature measurement, then the joint probability can determine how accurately the curvature can be measured, given the pixels observed on the boundary.

Finally, the term $p\,(\text{Child, Action} \,|\, \text{Parent})$ is defined as

$$\int_{\text{outcomes of action}} p \; (\text{Child, Outcome} \,|\, \text{Parent})$$

where $p\,(\text{Child, Outcome} \,|\, \text{Parent})$ is computed and stored a priori. For example, if the child is a pair of generalized cylinders, the action is an angular measurement between them, and the parent is a joint with known angular measure; then the above formula specifies the probability we would observe a given outcome (angle) given the true (model) angle. See [Binford 87] for an example of such a computation.

Now if a higher level Bayes node, e.g., a generalized cylinder, is not yet instantiated, but we wish to compute the value of actions at an instantiated lower level node, e.g., an observed edge of a generalized cylinder, then the probability, $p$ (generalized cylinder), must be estimated a priori for the recursive computation of value at the observed edge node. We take these priors to be the task-based likelihood that given objects are present in a scenario. For example, in an assembly line application, based on the current manufacturing task, we have an a priori notion of what parts to expect on the line.

Control in the influence diagram is effected by the top-level loop of: execute actions, accrue probabilities, compute values, maximize utility, and select actions. A version of this algorithm in terms of the Bayes-net is given in Figure 3-1.

Execution of actions can occur on multiple machines in a distributed environment. Results are summarized and returned asynchronously to the Bayes net. We have structured the model space, and therefore the Bayes net, such that the assumptions of Pearl's algorithm [Pearl 86] are fulfilled. This allows asynchronous updating and propagation of probabilities, throughout the net. Because no decision nodes are between Bayes nodes, Pearl's algorithm applies over the subsets of the influence diagram that are connected Bayes-nets. Note that this structuring of the influence diagram, see Figure 2-4, was necessary to permit a control structure in which probability accrual, and decision making are separable operations.

```
Until ((time exceeded or (termination condition achieved))
    For each instantiated Bayes node
        Get list of possible actions from node
        Evaluate value of each action
    End for
    Until (all processors allocated or all actions selected)
        Maximize expected value constrained by total processing time
        Allocate selected actions over available processors
    End Until
        Until ((a node's probability ratio exceed's threshold)
            or (all k actions return values))
            propagate evidential returns over Bayes net
            update values at node
        End Until
End Until
```

Figure 3-1: Control Algorithm

## 4. EXAMPLE

The following example presents the use of utility-based control to drive the recognition of military units from aerial imagery. The aerial imagery used is assumed to be relatively low resolution so that individual vehicles are difficult to identify due to their small size and a high false alarm rate. As a result additional contexual forms of evidence are used to recognize the military forces. The acquired evidence is matched against known military force models in order to determine its support. The force models resident in the system are shown in Figure 2-2. The recognition system normally commences processing by generating hypotheses for the coarse models and proceeds by refining them and using them to generate higher-level hypotheses. The Bayes net is then used to group conflicting hypothesis configurations and to propagate beliefs throughout the hypothesis space.

In this example we are attempting to confirm the presence of a Brigade in the boxed region in the upper portion of the map shown in Figure 4-1a. The system is initialized by locating possible vehicle detections in the available imagery. These detections are then clustered into Company-size Unit hypotheses based on coarse parameters in the model database such as inter-vehicle distance, number of vehicles in a unit, and maximum extent of a unit. By initializing the Bayes net with Company-size Units and not individual vehicles we achieve a large reduction in

a.                                                       b.

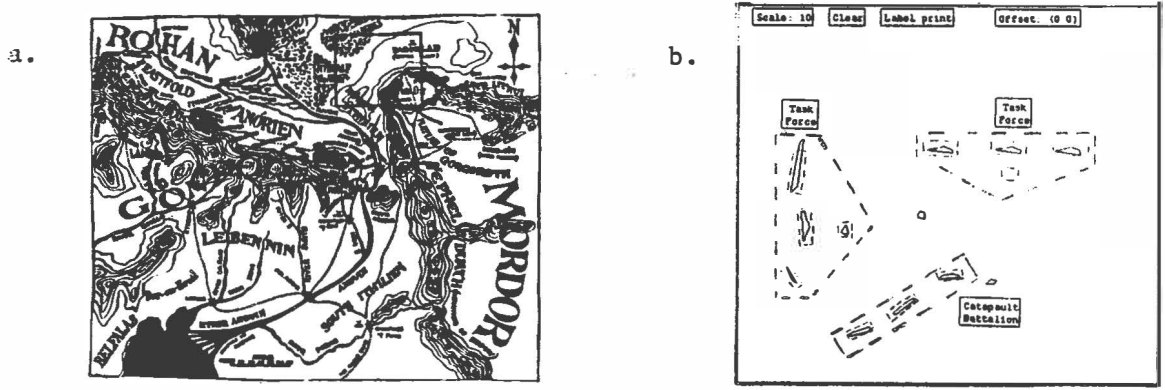

Figure 4-1: a. Area of Interest  b. Initial Hypotheses (Solid Lines) and
Actual Ground Truth (Dashed Lines)

combinatorics. The initial cluster units are shown in Figure 4-1b.

After initialization the system progresses by performing any of the following actions on the appropriate Bayes nodes:

- Refine a Bayes net hierarchy by using the more detailed force type model description (Refine-type).

- Refine a Bayes net hierarchy by using a more detailed formation description (Refine-formation).

- Search for matches among lower-level force hypotheses in order to generate higher-level force hypotheses (Search).

- Attach terrain evidence to a Bayes node by examining the support the underlying terrain provides for the given force (Terrain-support).

- Attach classification evidence to a Bayes leaf node indicating the support for the given force type obtained from high resolution sensors--an accurate process that is normally expensive to perform (Classification).

At each step, the system generates the available actions and computes their utility based on value and cost models derived from previous system performance. Optimal actions are then selected by maximizing the expected value of the actions that can be executed in a given time step using the knapsack approximation algorithm. Table 4-1 lists the actions selected by the system at each step.

The initial Bayes net after generation of Company-size Unit hypotheses is shown in Figure 4-2a. The attached probabilities are the ones obtained from the detection likelihoods and clustering matches. After the actions in step 1 were performed, the Bayes net contained refined hypotheses for Team, Task Force Headquarters, and Catapault Battery. These refined hypotheses are shown in Figure 4-2b. The next iteration executed actions supplying terrain and classification support for the four Bayes nodes. The resultant probabilities are shown in Figure 4-3a. The hypotheses with high belief are depicted graphically in Figure 4-3b. The third iteration step searched for matches among the likely hypotheses and generated Task Force and Catapault Battalion hypotheses. These are shown in Figures 4-4a and 4-4b. The fourth step directed the system to match the Task Force and Catapault Battalion hypotheses into Brigade hypotheses. The resultant Bayes net is shown in Figure 4-5 and the locations of the resultant hypotheses are shown

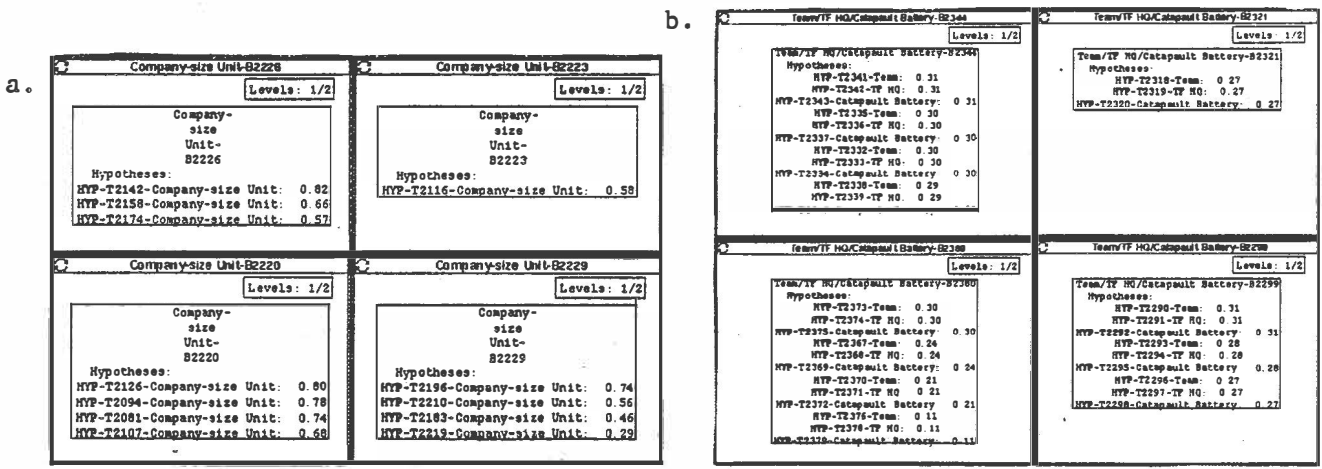

Figure 4-2: a. root nodes of the initialized Bayes net
b. root nodes of the Bayes net after the first step

251

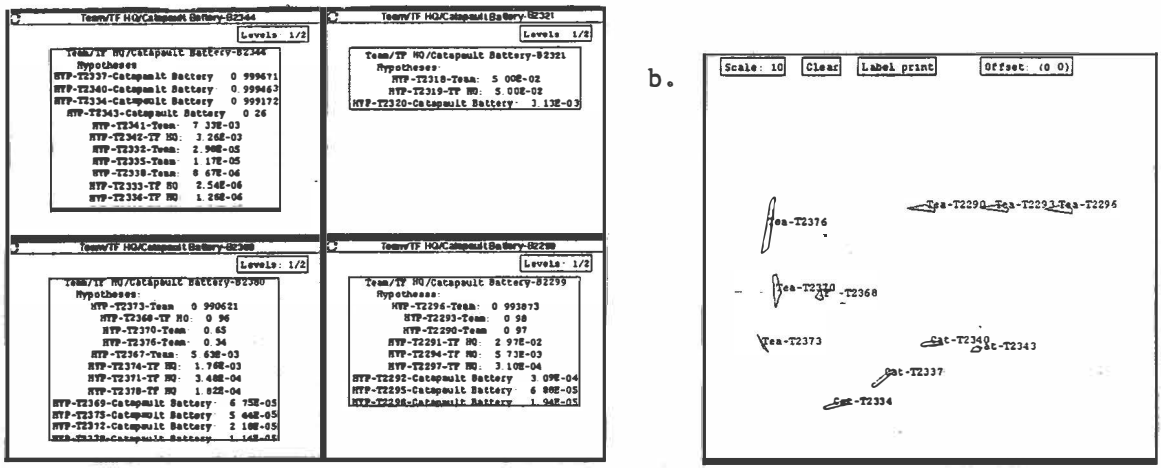

Figure 4-3: a. root nodes of the Bayes net after the second step
b. locations of likely hypotheses after the second step

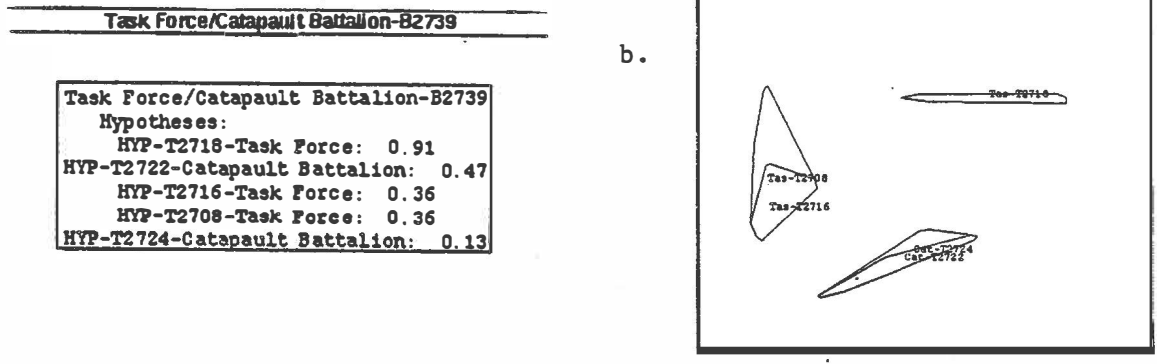

Figure 4-4: a. root node of the Bayes net after the third step
b. locations of likely hypotheses after the third step

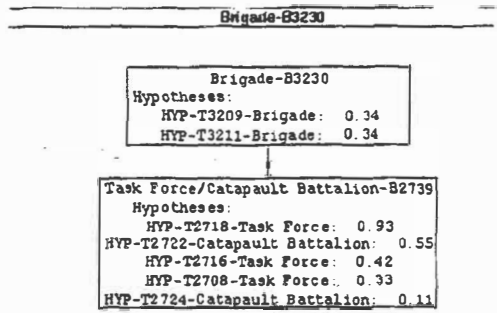

```
                    Brigade-B3230

         Brigade-B3230
    Hypotheses:
        HYP-T3209-Brigade:  0.34
        HYP-T3211-Brigade:  0.34

  Task Force/Catapault Battalion-B2739
    Hypotheses:
        HYP-T2718-Task Force:  0.93
  HYP-T2722-Catapault Battalion:  0.55
        HYP-T2716-Task Force:  0.42
        HYP-T2708-Task Force:  0.33
  HYP-T2724-Catapault Battalion:  0.11
```

Figure 4-5:    Bayes net after the fourth step

in Figures 4-6a and 4-6b. The fifth step attached terrain support to the Brigade hypotheses, a process that resulted in the Brigade hypotheses receiving high support (Figure 4-7). As a result the system reported that a Brigade most likely exists in the area (since the two hypotheses conflict, the belief that a Brigade is present is .99) and its exact location is given by the likelier hypothesis.

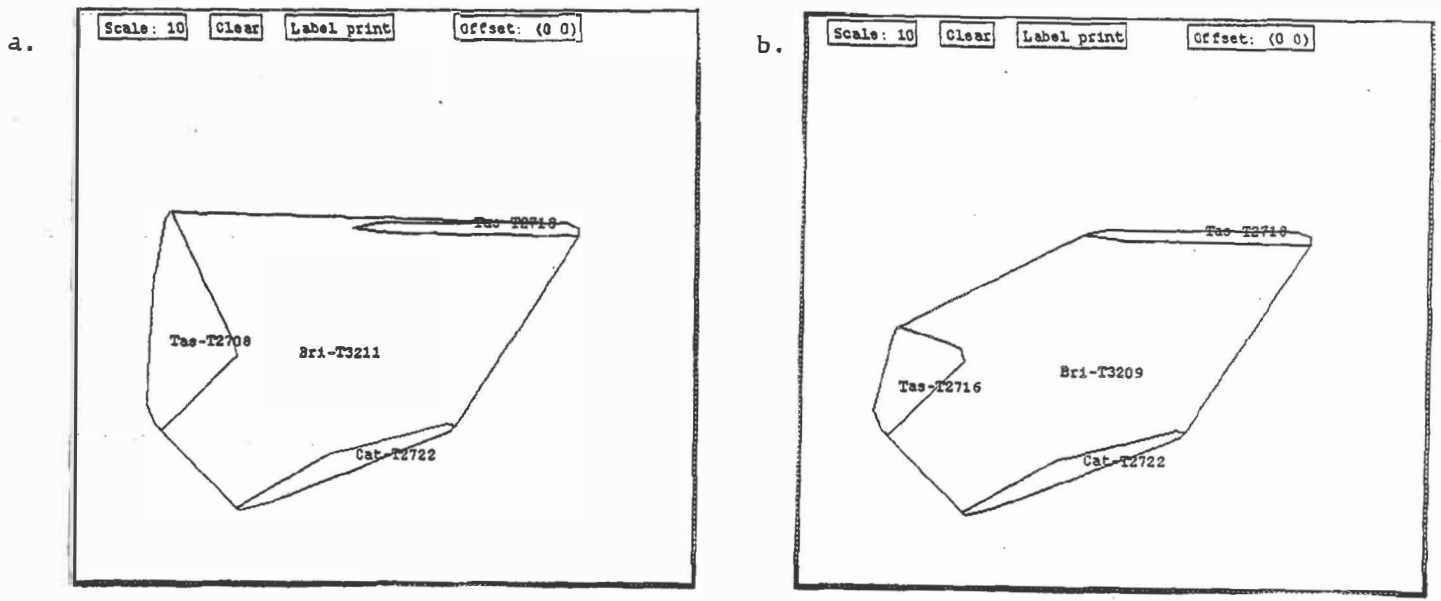

Figure 4-6: a. location of first Brigade hypothesis
b. location of second Brigade hypothesis

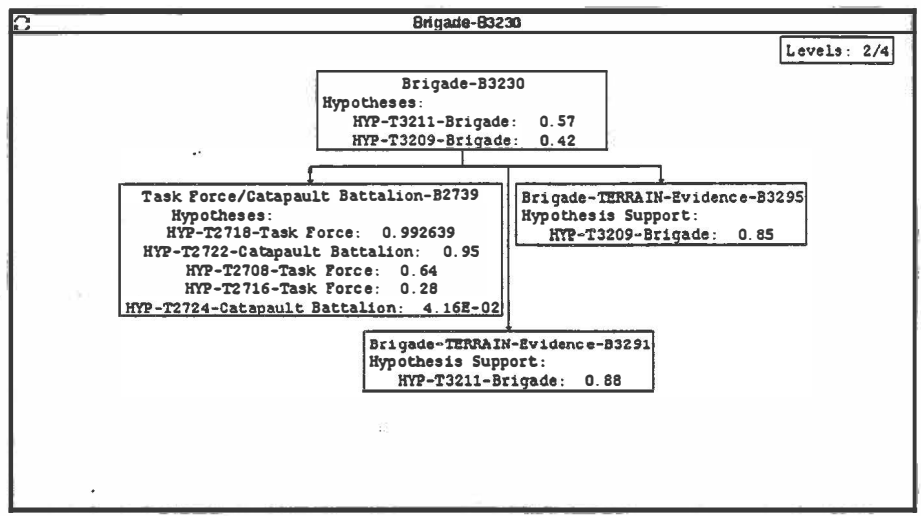

Figure 4-7: Bayes net after the fifth step

Table 4-1: Actions Selected by the System at Each Step

| Step | Selected Actions | Value | Cost |
|---|---|---|---|
| 1 | (REFINE-TYPE of(<B2226> <B2223> <B2220> <B2229>)) | 11522 | 1600 |
| | (SEARCH for matches of hypotheses in (<B2226> <B2223> <B2220> <B2229>)) | 5761 | 842 |
| | (TERRAIN-SUPPORT for <B2220>) | 1125 | 820 |
| | (TERRAIN-SUPPORT for <B2226>) | 769 | 820 |
| | (TERRAIN-SUPPORT for <B2229>) | 769 | 820 |
| | (TERRAIN-SUPPORT for <B2223>) | 217 | 820 |
| 2 | (CLASSIFICATION-SUPPORT for <B2344>) | 619 | 1320 |
| | (TERRAIN-SUPPORT for <B2344>) | 326 | 820 |
| | (CLASSIFICATION-SUPPORT for <B2299>) | 445 | 1320 |
| | (CLASSIFICATION-SUPPORT for <B2380>) | 441 | 1320 |
| | (TERRAIN-SUPPORT for <B2299>) | 234 | 820 |
| | (TERRAIN-SUPPORT for <B2380>) | 232 | 820 |
| | (CLASSIFICATION-SUPPORT for <B2321>) | 139 | 1320 |
| | (TERRAIN-SUPPORT for <B2321>) | 73 | 820 |
| 3 | (SEARCH for matches of hypotheses in (<B2344> <B2321> <B2380> <B2299>)) | 2359 | 841 |
| 4 | (SEARCH for matches of hypotheses in (<B2739>)) | 1672 | 832 |
| | (REFINE-FORMATION of <B2739>) | 1060 | 8100 |
| | (SEARCH for matches of hypotheses in (<B2321>)) | 20 | 801 |
| 5 | (TERRAIN-SUPPORT for <B2739>) | 281 | 2300 |
| | (TERRAIN-SUPPORT for <B3230>) | 252 | 2300 |

## 5. CONCLUSIONS

We have developed a methodology for vision system control based on utility theory applied to model-based Bayesian inference. We have implemented this methodology in ADRIES, a radar surveillance system and are implementing it in SUCCESSOR, a system for computer vision of industrial parts in optical imagery. Many technical innovations were developed including:

- Representation of control and inference in a cognitively tractable model promoting clean and efficient system designs.

- Separation of decision making from evidence accrual.

- Dynamic instantiation of Bayes nets and influence diagrams.

- Hierarchical value computation achieved by assigning values only at the top model-level.

- Handling real world problems.

## 6. ACKNOWLEDGEMENTS

This work was supported by the Advanced Digital Radar Imagery Exploitation Systems (ADRIES) project sponsored by the Defense Advanced Research Projects Agency (DARPA) and the U.S. Army Engineer Topographic Laboratories (USAETL) under U.S. Government Contract No. DACA76-86-C-0010, and by the Knowledge-Based Vision Techniques (KBV) project sponsored by USAETL and DARPA under U.S. Government Contract No. DACA76-85-C-0005. Thanks are due to Carol Etheridge for document support.